\documentclass[letterpaper, 10 pt, conference]{ieeeconf}  

\IEEEoverridecommandlockouts                              
\usepackage[noend]{algpseudocode}
\usepackage[font={small,it}]{caption}
\usepackage{float}
\usepackage{graphicx}
\usepackage{epsfig}
\usepackage{amsmath}
\usepackage{amssymb} 
\usepackage{longtable}
\usepackage{algorithm}
\usepackage{rotating}
\usepackage{multirow}
\usepackage{array}
\usepackage{graphicx}
\usepackage{mathrsfs}
\usepackage{verbatim}
\usepackage{color}
\usepackage{epstopdf}
\usepackage{url}
\usepackage{graphicx}
\usepackage{mathrsfs}
\usepackage{verbatim}
\usepackage{color}
\usepackage[noadjust]{cite}
\usepackage{epstopdf}
\usepackage{url}
\usepackage{subfig}

\usepackage{graphicx}
\usepackage{epsfig}
\usepackage{amsmath}
\usepackage{amssymb} 
\usepackage{longtable}
\usepackage{rotating}
\usepackage{multirow}
\usepackage{array}
\usepackage{graphicx}
\usepackage{mathrsfs}
\usepackage{verbatim}
\usepackage{color}
\usepackage{epstopdf}
\usepackage{url}
\usepackage{graphicx}
\usepackage{mathrsfs}
\usepackage{verbatim}
\usepackage{soul}
\usepackage{epstopdf}
\usepackage{url}

\usepackage{mwe}

\usepackage{pgfplots,tikz-3dplot}
\usepackage{amsmath} 
\usepackage{amssymb}  
\usepackage{bbm}
\usepackage{color}
\usepackage{graphics}

\usepackage{epsfig}
\usepackage{epstopdf}
\usepackage{amsfonts}
\usepackage{mathtools}
\usepackage{bm}
\usepackage{fancyhdr}
\usepackage{framed}
\usepackage{float}
\usepackage{verbatim}
\usepackage{tikz}
\usepackage{relsize}
\usepackage{mathrsfs}
\usepackage{todonotes}
\usepackage{pgfplots}
\usepackage{grffile}
\usepackage{hyperref}

\usetikzlibrary{arrows}

\usepackage{todonotes}

\newcommand{\diag}{\mathop{\mathrm{diag}}}

\newcommand{\eref}[1]{(\ref{#1})}
\newcommand{\zerom}[1]{\textbf{O}_{#1}}

\newtheorem{remark}{\bf{Remark}}
\newcommand{\bs}{\boldsymbol}

\newcommand{\ie}{i.e., }
\newcommand{\realR}[1]{\mathbb{R}^{#1}}
\newcolumntype{C}[1]{>{\centering\let\newline\\\arraybackslash}m{#1}}

\title{\Large \bf A Robust Model Predictive Control Approach for Autonomous Underwater Vehicles Operating in a Constrained workspace 	}

\author{Shahab Heshmati-alamdari, George C. Karras, Panos Marantos, and Kostas J. Kyriakopoulos
\thanks{The authors are with the Control Systems Lab, Department of Mechanical Engineering, National Technical University of Athens, 9 Heroon Polytechniou Street, Zografou 15780, Greece. Email: {\tt\small \{shahab,karrasg,marantos,kkyria@mail.ntua.gr\}} }%
}

\begin{document}
\maketitle \thispagestyle{empty} \pagestyle{empty}

\begin{abstract} 
This paper presents a novel Nonlinear Model Predictive Control (NMPC) scheme for underwater robotic vehicles operating in a constrained workspace including static obstacles. The purpose of the controller is to guide the vehicle towards specific way points. Various limitations such as: obstacles, workspace boundary, thruster saturation and predefined desired upper bound of the vehicle velocity are captured as state and input constraints and are guaranteed during the control design. The proposed scheme incorporates the full dynamics of the vehicle in which the ocean currents are also involved. Hence, the control inputs calculated by the proposed scheme are formulated in a way that the vehicle will exploit the ocean currents, when these are in favor of the way-point tracking mission which results in reduced energy consumption by the thrusters. The performance of the proposed control strategy is experimentally verified using a $4$ Degrees of Freedom (DoF) underwater robotic vehicle inside a constrained test tank with obstacles. 
\end{abstract}

\section{Introduction}\label{Sec:Introduction}

During the last decades, underwater robotic vehicles have been widely used in a variety of marine activities. Applications such as monitoring and inspection, surveillance of underwater facilities, oceanography, search and rescue, are indicative examples of applications that require the underwater robot to operate under various constraints and increased level of autonomy, in terms of energy consumption and endurance  \cite{griffiths2002technology,Fossen2011}. Thus, newest research directions point towards the development of motion control strategies that are able to handle complex missions with reduced energy requirements \cite{Zeng2015303}.

The energy consumption of an underwater vehicle can be distinguished in two parts: i) the hotel load which is defined as the energy consumption due to the on-board computers, processing effort, instrumentation and communication devices ii) the energy used by the propulsion system (e.g thrusters) \cite{Zeng2015303}. The hotel load reduction can be achieved by employing low power devices and lean algorithms that do not require significant processing effort. On the other hand, the optimization of the thrust energy consumption, is mainly a path planning problem where the vehicle must reach the desired goal in an energy optimal manner. 

Energy minimization via mission planning has been studied in the past \cite{Huynh20151144,Alvarez2004418,Garau20095} and it is still an active area of research for underwater robotics. The importance of utilizing ocean currents in underwater vehicle operation was emphasized in \cite{Alvarez2004418}, where a genetic algorithm  planner was proposed for the design of a path with minimum energy requirements. An energy efficient framework was also proposed in \cite{Garau20095}, where the authors consider quasi-static ocean current information and constant thrust power. In \cite{Pêtrès2007331}, an $A^*$ search is used in order to design a continuous path where the ocean currents were incorporated as quadratic drag force terms.

\begin{figure}[t!]
	\begin{center}
		\includegraphics[width=3.3in]{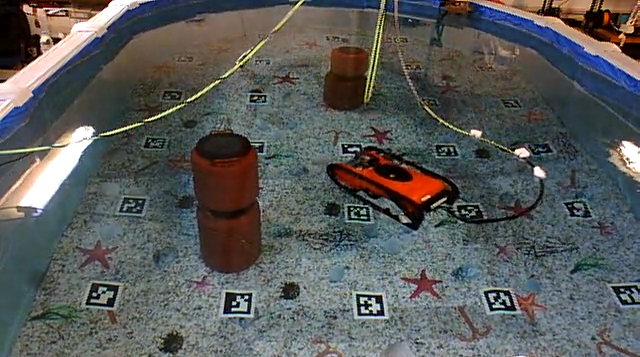}
	\end{center}
	\caption{Experimental setup and problem formulation: the purpose of the controller is to guide the vehicle towards desired way points inside a constrained workspace including sparse obstacles.  \vspace{-6mm} }\label{new_fig1}
\end{figure}


Moreover, the motion control of underwater vehicles is a highly nonlinear problem, where multiple input and state constraints are imposed to the system. Several linear and nonlinear motion control techniques for underwater vehicles can be found in literature. However, input (generalized body forces/torques or thrust)  and state (3D obstacles, velocities) constraints  are not always considered. Nonlinear Model Predictive Control (NMPC) \cite{allgower2004}, is an ideal approach for complex underwater missions, as it is able to combine motion planning, obstacle avoidance and workspace restrictions, while handling efficiently input and state constraints. 

In \cite{Huynh20151144}, the authors propose an MPC scheme in order to design an energy efficient path for a glider,  by minimizing a cost function based on the consumed energy. However, only the kinematic model of the vehicle is considered, without any disturbances or noise of the ocean current profile. In order to optimize sawtooth paths for an Autonomous Underwater Vehicle (AUV), an MPC scheme with a least squares cost function is presented in \cite{Medagoda2012}. Interesting results including estimated ocean wave profiles into an NMPC scheme, with an emphasis on real-time execution are presented in \cite{Fernandez201788}. However, the effect of noise and disturbance are not theoretically considered, but instead are presented through simulation testing. An MPC scheme with reduced dynamic model is presented in \cite{Jagtap2016772}, where in order to avoid computational complexity, simplified linear models were considered for the vertical and horizontal control of the vehicle. In the aforementioned works the validation of the proposed strategies was conducted via simulation tests. An experimental validation of a visually aided NMPC scheme for an underwater robotic system was presented in \cite{Heshmati-Alamdari20143826}, where simple kinematic equations of the system was considered.

\subsection{Contribution}
In this work, a novel NMPC scheme for underwater robotic vehicles is presented. The purpose of the controller is to guide the vehicle towards specific way points (See Fig.\ref{new_fig1}). Various constraints such as: sparse obstacles, workspace boundaries, control input saturation as well as predefined upper bound of the vehicle velocity (requirements for several underwater tasks such as seabed inspection scenario, mosaicking) are considered during the control design. The proposed scheme incorporates the full dynamics of the vehicle in which the ocean currents are also involved. The controller is designed in order to find optimal thrusts required for minimizing the way point tracking error. Moreover, the control inputs calculated by the proposed approach are formulated in a way that the vehicle will exploit the ocean currents, when these are in favor of the way-point tracking mission, which results in reduced energy consumption by the thrusters. The performance of the proposed control strategy is experimentally verified using a $4$ DoF underwater robotic vehicle inside a constrained test tank with obstacles. To the best of the authors knowledge, this is the first time where a NMPC scheme which incorporates the full dynamics of the vehicle is experimentally verified in a constrained workspace including sparse obstacles. 
\section{PRELIMINARIES}\label{Sec:preliminaries}
\subsection{Notation}
In this work, the vectors are denoted with lower bold letters whereas the matrices by capital bold letters.  We define as $\mathcal{B}(\bs{c},r)=\{\bs{x} \in \mathbb{R}^3: \|\bs{x} -\bs{c} \| \le r\}$ the closed sphere with radius $r$ and center $\bs{c}$. For a given set $A \subset \mathbb{R}^n$ we define as $\text{cl}(A), \text{int}(A)$ and $\partial S = \text{cl}(A) \backslash \text{int}(A)$ its closure, interior and boundary, respectively. 
\subsection{Mathematical Modeling}
\label{preliminaries}
\begin{figure}
	\includegraphics[scale=0.69]{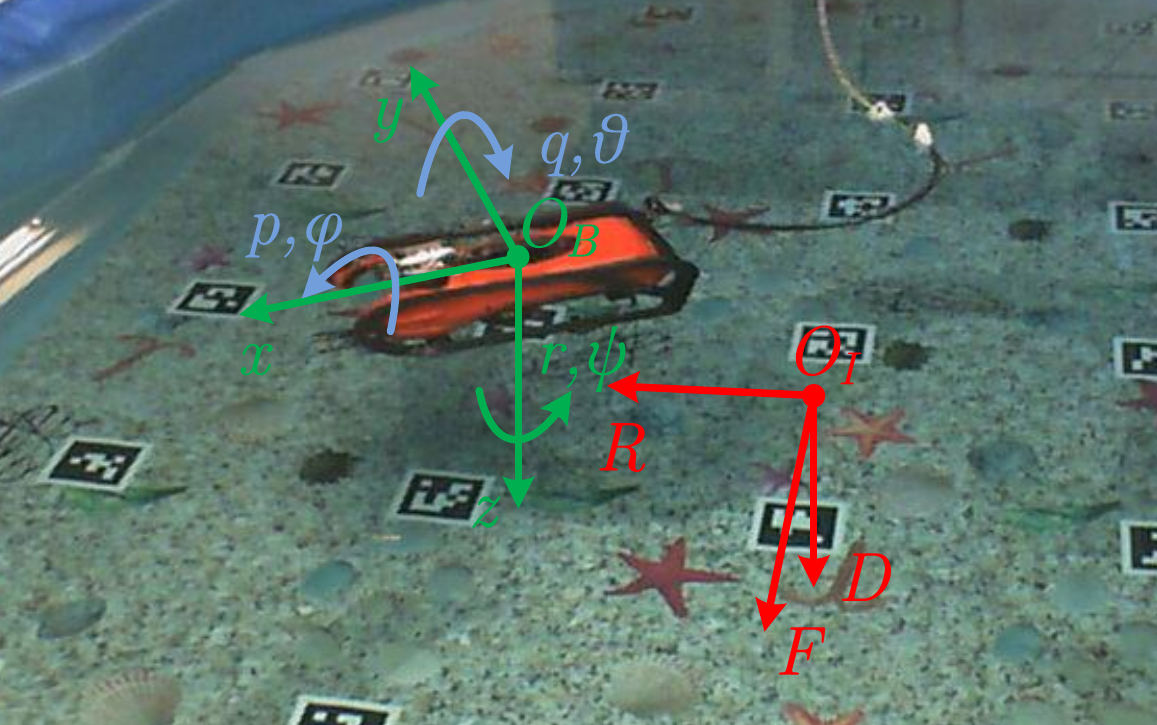}
	\centering
	\caption{Seabotix LBV150 ROV. The inertial frame ($\mathcal{I}$) and body-frame ($\mathcal{V}$) are indicated in red and green color  respectively. The under-actuated DoFs are also depicted in blue color.\vspace{0mm}}
	\label{fig:LBV_frames}
\end{figure}
The prior step before analyze the proposed methodology is the presentation of the preliminary aspects of the modeling of underwater vehicles. Firstly, let us define a common body-fixed frame $\mathcal{V}=\{e_x,e_y,e_z\}$ attached to the vehicle center of gravity, as well as the inertial frame $\mathcal{I}=\{e_F,e_R,e_D\}$ as shown in Fig-\ref{fig:LBV_frames}. The pose vector of the vehicle with respect to (w.r.t) the inertial frame $\mathcal{I}$ is denoted by $\boldsymbol{\eta}=\left[\boldsymbol{\eta^{T}_1}~\boldsymbol{\eta^{T}_2} \right]^T\in\realR{6}$ including the position (\ie $\boldsymbol{\eta_1}=\left[x~y~z\right]^T$) and orientation (\ie $\boldsymbol{\eta_2}=\left[\phi~\theta~\psi\right]^T$)
vectors. The $\boldsymbol{v}=\left[\boldsymbol{v^{T}_1}~\boldsymbol{v^{T}_2} \right]^T\in\realR{6}$ is the velocity vector of the vehicle expressed in fixed-body frame $\mathcal{V}$ and includes the linear (\ie $\boldsymbol{v_1}=\left[u~v~w\right]^T$) and angular (\ie $\boldsymbol{v_2}=\left[p~q~r\right]^T$) velocity vectors.  In this work, we consider that the vehicle operates under the influence of bounded irrotational ocean currents w.r.t the inertial frame $\mathcal{I}$. An estimation of the ocean currents can be achieved by employing the data obtained from Naval Coastal Ocean Model (NCOM) \cite{NCOM} and Regional Ocean Model Systems (ROMS)\cite{Smith20101475}. However, an estimation of the ocean current could be achieved locally using an appropriate estimator \cite{Aguiar20071092}. Thus, in the following analysis, we consider the effect of ocean currents during the control design. In this work, the bounded irrotational ocean current velocities w.r.t the inertial frame $\mathcal{I}$ is denoted by $\bs{v}^{\mathcal{I}}_c=[(\bs{v}^{\mathcal{I}}_{c_1})^T,\bs{0}_{1\times 3}]^T\in\mathbb{R}^6$ with $\bs{v}^{\mathcal{I}}_{c_1}=[u_{c}^{\mathcal{I}},v_{c}^{\mathcal{I}},w_{c}^{\mathcal{I}}]^T$ to be the vector of linear velocity terms. Therefore, we can define the vehicle velocity vector relative to the water expressed in body frame $\mathcal{V}$ as
\begin{align}
\bs{v}_r=\bs{v}-\bs{v}_c\label{vel_rel}
\end{align}
Notice that the vector $\bs{v}_c=[u_c,v_c,w_c,\bs{0}_{1\times 3}]^T$ indicates the expression of the ocean currents with respect to the body frame $\mathcal{V}$. Without loss of generality, according to the standard underwater vehicles' modeling properties \cite{Fossen2}, assuming that the current velocity is slowly varying with respect to the inertial frame (e.g, $\frac{\partial\bs{v}^{\mathcal{I}}_c}{\partial t}\approxeq 0$), and the vehicle is operating at relative low speeds, the dynamic equations of the vehicle can be given as \cite[eq:3.110-3.116]{Fossen2}:%
\begin{subequations}\label{eq:AUV_model}
	\begin{gather}%
	{\boldsymbol{\dot{\eta}}}=\boldsymbol{J}\left( \boldsymbol{ \eta}\right)  \boldsymbol{v}_r+\bs{v}^{\mathcal{I}}_c\\
	\boldsymbol{M}{\boldsymbol{\dot{v}}}_r\!+\! \boldsymbol{C}\left(  \boldsymbol{v}_r\right)  \boldsymbol{v}_r\!+\! \boldsymbol{D}(\bs{v}_r) \boldsymbol{v}_r\! +\! \boldsymbol{g}\left(  \boldsymbol{\eta}\right)
	=\boldsymbol{\tau }_\mathcal{V}
	\end{gather}
\end{subequations}
where:
\begin{itemize}
	\item $\boldsymbol{\tau}_\mathcal{V}=\left[X,~Y,~Z,~K,~M,~N\right]^T\in\realR{6}$ is the total propulsion force/torque vector (\ie the body forces and torques generated by the thrusters) applied on the vehicle and expressed in $\mathcal{V}$;
	
	\item $\boldsymbol{M}=\boldsymbol{M}_{RB}+\boldsymbol{M}_{A}$, where $\boldsymbol{M}_{RB}\in\realR{6\times6}$ and $\boldsymbol{M}_{A}\in\realR{6\times6}$  are the inertia matrix for the rigid body and added mass respectively;
	
	\item $\boldsymbol{C}\left(  \boldsymbol{v}_r\right) = \boldsymbol{C}_{RB}\left(  \boldsymbol{v}_r\right)  +\boldsymbol{C}_{A}\left(  \boldsymbol{v}_r\right)
	~$, where $\boldsymbol{C}_{RB}\left(  \boldsymbol{v}_r\right)\in\realR{6\times6}$ and $\boldsymbol{C}_{A}\left(  \boldsymbol{v}_r\right)\in\realR{6\times6}$ are the coriolis and centripetal matrix for the rigid body and added mass respectively;
	
	\item $\boldsymbol{D}\left(\boldsymbol{v}_r\right)  = \boldsymbol{D}_{quad}\left(  \boldsymbol{v}_r\right)  +\boldsymbol{D}_{lin}\left(  \boldsymbol{v}_r\right)
	~$, where $\boldsymbol{D}_{quad}\left(  \boldsymbol{v}_r\right)\in\realR{6\times6}$ and $\boldsymbol{D}_{lin}\left(  \boldsymbol{v}\right)\in\realR{6\times6}$ are the quadratic and linear drag matrix respectively;
	
	\item $\boldsymbol{g}\left(\boldsymbol{\eta}\right)\in\realR{6}$ is the hydrostatic restoring force vector;
	
	\item $\boldsymbol{J}\left(  \boldsymbol{\eta}\right)=\left[\begin{array}{cc}
	\boldsymbol{J}_1\left(\boldsymbol{\eta_2}\right) & \zerom{3\times 3}\\
	\zerom{3\times 3} & \boldsymbol{J}_2\left(\boldsymbol{\eta_2}\right)
	\end{array}\right]$ is the Jacobian matrix transforming the
	velocities from the body-fixed ($\mathcal{V}$) to the inertial ($\mathcal{I}$) frame, in which $\boldsymbol{J}_1\left(\boldsymbol{\eta_2}\right)\in SO(3)$ is the well known rotation matrix and $\boldsymbol{J}_2\left(\boldsymbol{\eta_2}\right)\in\realR{3\times3}$ denotes the lumped transformation matrix;
\end{itemize}
Notice that the transformation from ocean current velocity defined in the inertial frame $\mathcal{I}$ (i.e., $\bs{v}^{\mathcal{I}}_c$) into body-fixed one (i.e., $\bs{v}_c$) is achieved using the transposed rotation matrix i.e, $\bs{v}_c=\boldsymbol{J}^T(\boldsymbol{\eta}) \bs{v}^{\mathcal{I}}_c$ (See\cite{Fossen2}). In \eref{eq:AUV_model}, the total propulsion force/torque vector ($\boldsymbol{\tau}_\mathcal{V}$) is computed using the thruster allocator matrix which is formulated by the actuation geometry and properties of the underwater vehicle's thrusters. The vehicle used in this work is a $4$ DoF Seabotix LBV150. It is equipped with $4$ thrusters (\ie Port ($p_o$), Starboard ($s$), Vertical ($v_e$), Lateral ($l$)), which are effective in Surge ($X$), Sway ($Y$), Heave ($Z$) and Yaw ($N$) motion. Thus, we can define a new thrust vector ($\boldsymbol{\tau}=\left[\tau_{p_o},~\tau_s,~\tau_{v_e},~\tau_l\right]^T\in\realR{4}$) and the appropriate thruster allocator matrix ($\boldsymbol{T_A}\in\realR{4\times4}$) such as:
\begin{equation}
\label{eq:thruster_allocator}
\boldsymbol{\tau}_\mathcal{V}^{LBV}=\boldsymbol{T_A}\boldsymbol{\tau},
\end{equation}
where $\boldsymbol{\tau}_\mathcal{V}^{LBV}\left[X,~Y,~Z,~N\right]^T\in\realR{4}$.
\begin{remark}
In the vehicle used in this work, the angles \(\phi\), \(\theta\) and angular velocities \(p\) and \(q\) are negligible
	and we can consider them to be equal to zero. Thus, from now on, the $\bs{\eta}=[x,y,z,\psi]$ and $\bs{v}=[u,v,w,r]$. The vehicle is symmetric about \(x\) - \(z\) plane and close to symmetric about \(y\) - \(z\) plane. Therefore, we can safely assume that motions in
	heave, roll and pitch are decoupled \cite{Fossen2}. 
\end{remark}

\section{METHODOLOGY}\label{Sec:methodology}
In this Section we present in detail the methodologies proposed in order to guide the vehicle towards a set of way-points a set of way-points $\bs{\eta}^d_i,~i=\{1,\ldots,n\}$.
\subsection{Geometry of Workspace}
Consider an underwater vehicle which operates inside the workspace $\mathcal{W}\subset \mathbb{R}^3$ with boundaries  $\partial\mathcal{W}=\{ \bs{p}\in \mathbb{R}^3: \bs{p} \in \text{cl}(\mathcal{W}) \backslash \text{int}(\mathcal{W}) \}$ and scattered obstacles located within. Without any loss of the generality, the robot and the obstacles are all modeled by spheres (i.e., we
adopt the spherical world representation\cite{Koditschek1990412}). 
In this spirit, let $\mathcal{B}(\boldsymbol{\eta}_1, \bar{r})$ to be a closed ball that covers all the vehicle volume (main body and additional equipments). Moreover, the $\mathcal{M}$ statics obstacles within the workspace are defined as closed spheres described by $\pi_m=\mathcal{B}(\bs{p}_{\pi_m},r_{\pi_m}),~ m\in\{1,\ldots,\mathcal{M}\}$, where $\bs{p}_{\pi_m}\in \mathbb{R}^3$ is the center and the $r_{\pi_m}>0$ the radius of the obstacle $\pi_m$. Additionally, based on the property of spherical world\cite{Koditschek1990412}, for each pair of obstacles $m,m'\in\{1,\ldots,\mathcal{M} \}$ the following inequality holds:\vspace{0mm}
\begin{align}
|| \pi_{m}-\pi_{m'}|| > 2\bar{r} + r_{\pi_m}+r_{\pi_m'}
\label{obstacle_dis}
\end{align} 
which intuitively means that the obstacles $m$ and $m'$ are disjoint in such a way that the entire volume of the vehicle can pass through the free space between them. Therefore, there exists a feasible trajectory $\bs{\eta}(t)$ for the vehicle that connects the initial configuration $\bs{\eta}(t_0)$ with $\bs{\eta}^d$ such as: \vspace{0mm}
\begin{align*}
\mathcal{B}(\boldsymbol{\eta}_1(t), \bar{r})\cap \{ \mathcal{B}(\bs{p}_{\pi_m},r_{\pi_m}) \cup \partial \mathcal{W} \} = \emptyset,~  m \in\{1,\ldots,\mathcal{M}\}
\end{align*}
\subsection{Dynamical system}
Due to the aforementioned assumptions and following standard simplifications due to symmetries in the mass configuration \cite{Fossen2}, the dynamic equation \eqref{eq:AUV_model} for the vehicle under consideration, can be written in discrete-time form as:
\begin{gather}
\bs{x}_{k+1}\!=\!f(\bs{x}_k,\bs{\tau}_k) \Rightarrow \bs{x}_{k+1}\!=\!\bs{x}_k\!+\!\mathcal{A}(\bs{x}_k)dt+\mathcal{C}(\bs{\tau}_k)dt\label{eq3} \end{gather} where:{\small{\begin{gather*}
		\!\mathcal{A}(\bs{x}_k)\!=\!\!\begin{bmatrix}u_{r_k} c\psi_k-v_{r_k}s\psi_k+u_c^{\mathcal{I}} \\ u_{r_k}s\psi_k+v_{r_k}c\psi_k+v_c^{\mathcal{I}}\\ w_{r_k}+w_c^{\mathcal{I}}\\ r_{r_k}\\\frac{1}{m_{11}}(m_{22}v_{r_k} r_{r_k} +X_u u_{r_k}+ X_{u|u|}|u_{r_k}|u_{r_k})\\
		\frac{1}{m_{22}}(-m_{11}u_{r_k} r_{r_k}+Y_v v_{r_k}+Y_{v|v|}|v_{r_k}|v_{r_k})\\
		\frac{1}{m_{33}}(Z_w w_{r_k}+Z_{w|w|}|w_{r_k}|w_{r_k})\\
		\frac{1}{m_{44}} ((m_{11}\!-\!m_{22})u_{r_k} v_{r_k}\!\!+\!\!N_rr_{r_k}\!+\!N_{r|r|}|r_{r_k}|r_{r_k} )\!\end{bmatrix}
		,\vspace{+4mm}\\\mathcal{C}(\bs{\tau}_k)=\begin{bmatrix}
		\bs{0}_{4\times 1}\\ \bs{T_A}\bs{\tau}_k
		\end{bmatrix}\end{gather*}}}
with $c(\cdot)=\cos(\cdot),s(\cdot)=\sin(\cdot)$ and $\mathbf{x_k}=[\bs{\eta}_k^T,\bs{v}^T_{r_k}]^T=[x_k,~y_k,~z_k,~\psi_k,~u_{r_k},~v_{r_k},~w_{r_k},~r_{r_k}]^\top\in \mathbb{R}^8$ denotes the state vector at
the time-step $k$ which includes the position and orientation of the vehicle with respect to the inertial frame $\mathcal{I}$ and the relative linear and angular velocity of the vehicle with respect to the water. In addition, $m_{ii}, i=1,\ldots,4$ are the mass terms including added mass, $X_u,~Y_v,~Z_w,~N_r~<0$ are the linear drag terms, $X_{u|u|},~Y_{v|v|},~Z_{w|w|},~N_{r|r|}~<0$ are the quadratic drag terms, while $dt$ denotes the sampling period. The control input of the system is $\bs{\tau}_k=[\tau_{p_k},~\tau_{s_k},~\tau_{v_k},~\tau_{l_k}]^T \in \mathbb{R}^4$ consisting of the thrusters' forces.
\subsection{Constraints}
\textbf{State Constraints:}\\
In this work,  we consider that the robot must avoid the obstacles and the workspace boundaries (test tank). Moreover, for the needs of several common underwater tasks (e.g., seabed inspection, mosaicking), the vehicle is required to move with relatively low speeds with upper bound denoted by the velocity vector ${\bs{v}}_p = [u_p\;v_p\;w_p\;r_p]^\top$ . These requirements are captured by the state constraint set $X$ of the system, given by: \vspace{0mm}
\begin{equation}
\mathbf{x_k }\in {X} \subset \mathbb{R}^8 \label{eq4}
\end{equation}\vspace{0mm}
which is formed by the following constraints:
\begin{subequations}
	\begin{gather}
	u_p+v_p-|u_r+v_r| \geq 0 \label{eq5a}\\
	w_p-|w_r| \geq 0 \label{eq5b} \\
	r_p-|r_r| \geq 0  \label{eq5d}\\
	\mathcal{B}(\boldsymbol{\eta}_1(t), \bar{r}) \cap \{ \mathcal{B}(\bs{p}_{\pi_m},r_{\pi_m}) \cup \partial \mathcal{W} \} = \emptyset,\label{eq5f}\\
	\nonumber m \in\{1,\ldots,\mathcal{M}\}\end{gather}
\end{subequations}
\textbf{Input Constraints:}\\
It is well known that the forces generated by the thrusters. Thus, we define the control constraint set ${T}$ as follows:
\begin{equation}
\bs{\tau}_k = [\tau_{{p_o}_k},~\tau_{s_k},~\tau_{{v_e}_k},~\tau_{l_k}]^T\in T \subseteq \mathbb{R}^4
\label{eq6}
\end{equation}
These constraints are of the form $|\tau_{{p_o}_k}|\leq \bar{\tau}_{p_o}$ , $|\tau_{s_k}|\leq \bar{\tau}_{s}$,
$|\tau_{{v_e}_k}|\leq \bar{\tau}_{v_e}$ and  $|\tau_{l_k}|\leq \bar{\tau}_{l}$, therefore we get $\||\bs{\tau}_k || \leq \bar{T}$ where $\bar{T}=(\tau_{p_o}^2+\tau_{s}^2+\tau_{v_e}^2+\tau_{l}^2)^{\frac{1}{2}}$ and
$\bar{\tau}_{p_o},\bar{\tau}_{s},\bar{\tau}_{v_e},\bar{\tau}_{l} \in \mathbb{R}_{\geq 0}$. 

\subsection{Control Design}
The control objective is to guide the regions around the waypoints $i=\{1,\ldots,n\}$ that includes the desired state $^i\mathbf{x}^d\triangleq [(^i\bs{\eta}^d)^T,(^i\bs{v}^d_r)^T]^T=[^ix_d,^iy_d,^iz_d,^i\psi_d,^iu_d,^iv_d,^iw_d,^ir_d]^T \in X$, while respecting the state constraints \eqref{eq5a}-\eqref{eq5f} as well as the  input constraints \eqref{eq6}.
A predictive controller is employed in order to achieve this task. In particular, at a given time instant $k$, the {NMPC} is assigned to solve an Optimal Control Problem (OCP)
with respect to a control sequence $\bs{\tau}_f(k)\triangleq [\mathbf{\tau}(k|k),\mathbf{\tau}(k+1|k),\dots,\mathbf{\tau}(k+N-1|k)]$, for a
prediction horizon $N$. The {OCP} of the {NMPC} is given as
follows:
\begin{subequations}
	\begin{align}&\min_{\bs{\tau}_f(k)}J_N(\bs{x}_k,\bs{\tau}_f(k))=\label{12a}\\ &\min_{\bs{\tau}_f(k)}\sum_{j=0}^{N-1}F(\hat{\bs{x}}(k+j|k),\bs{\tau}(k+j|k))+E(\hat{\bs{x}}(k+N|k))\nonumber\\
	&\text{subject to:}\nonumber\\
	&\hat{\bs{x}}(k+j|k)\in X_j, \qquad \,\,\,\,\forall j=1,\ldots,N-1 ,\label{12b}\\
	&\bs{\tau}(k+j|k) \in T,  \qquad \forall j=0, \ldots, N-1,\label{12c}\\
	&\hat{\bs{x}}(k+N|k) \in \mathscr{E}_f \label{12d}\end{align}
\end{subequations}
where  $\mathscr{E}_f$ is the terminal set and $F$ and $E$ are the running  and terminal cost functions, respectively.  At time instant $k$, the solution of the OCP \eqref{12a}-\eqref{12d} is providing an optimal control sequence, denoted as:
\begin{align}
\bs{\tau}_f^*(k)= [\mathbf{\tau}(k|k), \mathbf{\tau}(k+1|k),\dots, \mathbf{\tau}(k+N-1|k)]\label{eq14}
\end{align}
\noindent where the first control vector (i.e., $\bs{\tau}(k|k)$) is applied to the system. Notice we use the double subscript notation for the predicted state of system \eqref{eq3} inside the OCP of the NMPC:
\begin{equation}
\hat{\bs{x}}(k+j|k)=f(\hat{\bs{x}}(k+j-1|k),\mathbf{\tau}(k+j-1|k))\label{eq:hat}
\end{equation}
where the vector $\hat{\bs{x}}(k+j|k)$ denotes the predicted state of the system
\eqref{eq3} at sampling time $k+j$ with $j \in \mathbb{Z}_{\geq 0}$. The predicted
state is based on the measurement of the state $\bs{x}_k$ of the system at sampling time $k$ (i.e., provided by onboard navigation system), while applying a sequence of control inputs
$[\mathbf{\tau}(k|k), \mathbf{\tau}(k+1|k),\dots, \mathbf{\tau}(k+j-1|k)]$. It holds that $\hat{\bs{x}}(k|k)\equiv\bs{x}_k$.
The cost function $F(\cdot)$, as well as the terminal cost
$E(\cdot)$, are both of quadratic form, i.e.,
$F(\hat{\bs{x}},\mathbf{\tau})=\hat{\bs{x}}^\top Q\hat{\bs{x}}+\mathbf{\tau}^\top R\mathbf{\tau}$ and
$E(\hat{\bs{x}})=\hat{\bs{x}}^\top P\hat{\bs{x}}$, respectively, with $P$, $Q$
and $R$ being positive definite matrices. Particularly we define
$Q=\diag\verb|{|q_1,\ldots,q_8\verb|}|$,
$R=\diag\verb|{|r_1,\ldots,r_4\verb|}|$ and
$P=\diag\verb|{|p_1,\ldots,p_8\verb|}|$.

\section{EXPERIMENTAL RESULTS}\label{Sec:results}

This section demonstrates the efficacy of the proposed motion control scheme via a real-time experiment employing a small underwater robotic vehicle. In particular, Subsection \ref{setup} introduces the experimental setup and Subsection \ref{results} presents the detailed results of experimental studies with the proposed controller.

\subsection{Setup}
\label{setup}
The real time experiment was carried out inside the \textit{NTUA, Control Systems Lab} test tank, with dimensions $5m\times3m\times1.5m$ (Fig. \ref{new_fig1}). The bottom of the tank is covered by a custom-made poster with various visual features and markers. These visual features are used by a proper state estimator in order to provide the state of the vehicle. Two cylindrical objects with known position and dimensions are placed inside the tank and considered as static obstacles. The vehicle used in this work is a $4$ DoFs Seabotix LBV, actuated in Surge, Sway, Heave and Yaw via a $4$ thruster set configuration.The vehicle is equipped with a down-looking Sony PlayStation Eye camera, with $640\times480$ pixels at $30$ frames per second (fps) enclosed in a waterproof housing. An underwater laser pointer projecting a green dot at the bottom of the test tank is rigidly attached on the vehicle with its axes aligned to the down-looking camera axis. The projection of this laser dot on the image plane of the down-looking camera is used in order to provide the depth measurement. The vehicle is also equipped with an $SBG\ IG-500A$ AHRS, delivering temperature-compensated 3D acceleration, angular velocity and orientation measurements at $100Hz$. The marker localization system is based on the $ArUco$ library \cite{Aruco2014}. 

The complete state vector of the vehicle ($3D$ position, orientation, velocity) is available via a sensor fusion and state estimation module based on the Complementary Filter notion presented in our previous results \cite{Marantos20161214}. The vehicle's dynamic parameters have been identified via a proper identification scheme. The analysis of the sensor fusion, state estimation and parameter identification algorithms are out of the scope of this paper and thus omitted. The software implementation of the proposed motion control scheme was conducted in C\texttt{++} and Python under the Robot Operating System (ROS) \cite{Quigley09}. 

\begin{figure}[t!]
	\begin{center}
		\includegraphics[width=3.0in]{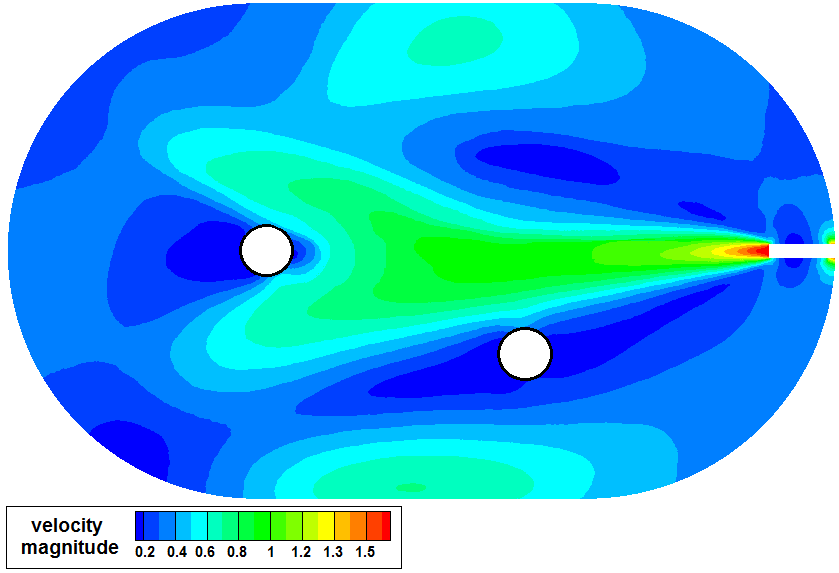}
	\end{center}
\vspace{0mm}\caption{Distribution of the flow field inside the experimental water tank as computed by the CFD software presented in \cite{Asouti2011232}.  \vspace{0mm} }\label{flow_fig}
\end{figure}

The disturbances in the form of water currents, were induced using a $BTD150$ thruster properly mounted inside the water tank. The generated flow field (i.e., assumed ocean current profile), was computed using a GPU-enabled Computational fluid Dynamics (CFD) software \cite{Asouti2011232} developed in the Parallel CFD and Optimization Unit of the school of Mechanical Engineering of NTUA. The flow field distribution inside the water tank is depicted in Fig-\ref{flow_fig}.

\subsection{Results}\label{results}
In order to prove the efficacy of the proposed controller a real time experiment is presented in this section. In this experiment, the objective is to follow a set of predefined waypoints while simultaneously avoid two static obstacles inside the workspace and respect the workspace (test tank) boundaries. The location and geometry of the obstacles are considered known. More specific, the position of the obstacles with respect to the Inertial Frame $\mathcal{I}$ in $x-y$ plane is given by: $
{\bf{x}}_{{\bf{obs}}_{\bf{1}} }  = \left[ {\begin{array}{*{30}c}
	{ - 0.625} & { - 0.625}  \\
	\end{array}} \right],\;{\bf{x}}_{{\bf{obs}}_{\bf{2}} }  = \left[ {\begin{array}{*{20}c}
	{0.9375} & 0  \\
	\end{array}} \right]$.

\begin{figure}
		\centering
	\vspace{0mm}	\includegraphics[width=3.3in]{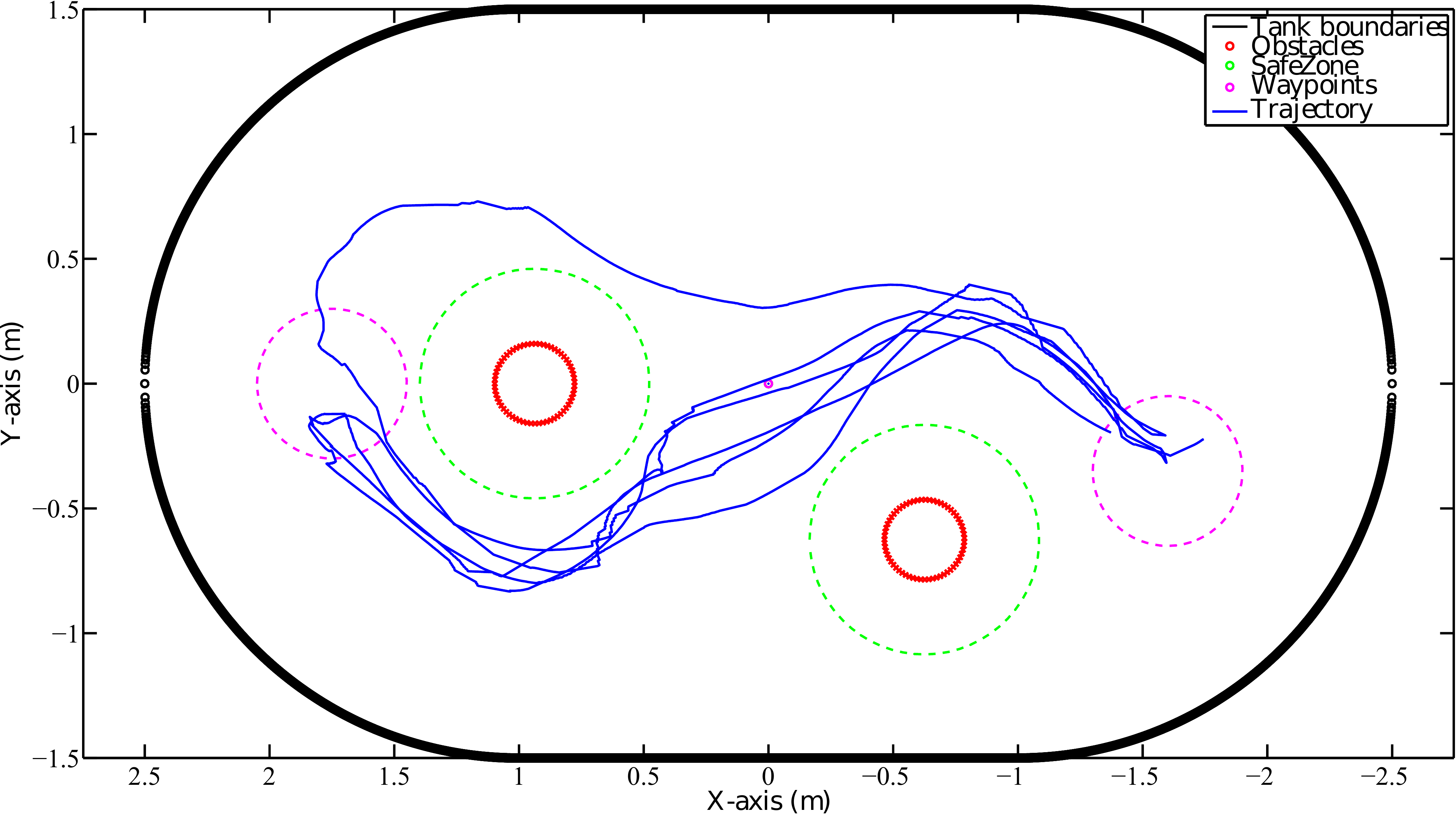}\vspace{0mm}
	\caption{2 WP tracking scenario: Vehicle trajectory in horizontal plane\vspace{0mm} }\label{2wpxytrajectory}
\end{figure}

The state constraints of the \eqref{eq5a}-\eqref{eq5f} which must be satisfied during the experimental operation are analytically formulated as follows: i) The obstacles are cylinders (See Fig.\ref{new_fig1}) with radius $r_{\pi_i} =0.16m,~i=\{1,2\}$ and are modeled together with the workspace boundaries according to the spherical world representations as consecutive spheres.  ii) the radius of the sphere $\mathcal{B}(\boldsymbol{\eta}_1, \bar{r})$ which covers all the vehicle volume (i.e., main body and additional equipment) is defined as $\bar{r} = 0.3m$. However, for the clarity of presentation, we depict it as a safe zone around the obstacles where the vehicle center $\boldsymbol{\eta}_1$ (denoted by blue line See Fig.\ref{2wpxytrajectory}) should not violated it. iii) the vertical position must be between $0 < z < 1.2\;m$, iv) the vehicle's body velocity norm of \eqref{eq5a} $|u_r+v_r|$ (planar motion) must not exceed $0.5 m/s$. v) heave velocity must be retained between  $-0.25 < w_r < 0.25\;m/s$. vi) yaw velocity must be retained between $-1 < r_r < 1\;rad/s$. 

Moreover, each of the four thrusters must obey the following input constraint: $- 12 < \tau_{i} < 12N,~i = \{p_o,~s,~v_e,~l\}$. The state and input constraints in the following figures are depicted in red dashed lines were applicable. At this point we should mention that the mission is considered as successful only if the vehicle performs the waypoint tracking three consecutive times. Thus, the repeatability of the proposed scheme is proved. In all times the vehicle is under the influence of the water currents depicted in Fig \ref{flow_fig}. 

\begin{figure}
		\centering
	\vspace{0mm}	\includegraphics[width=3.3in]{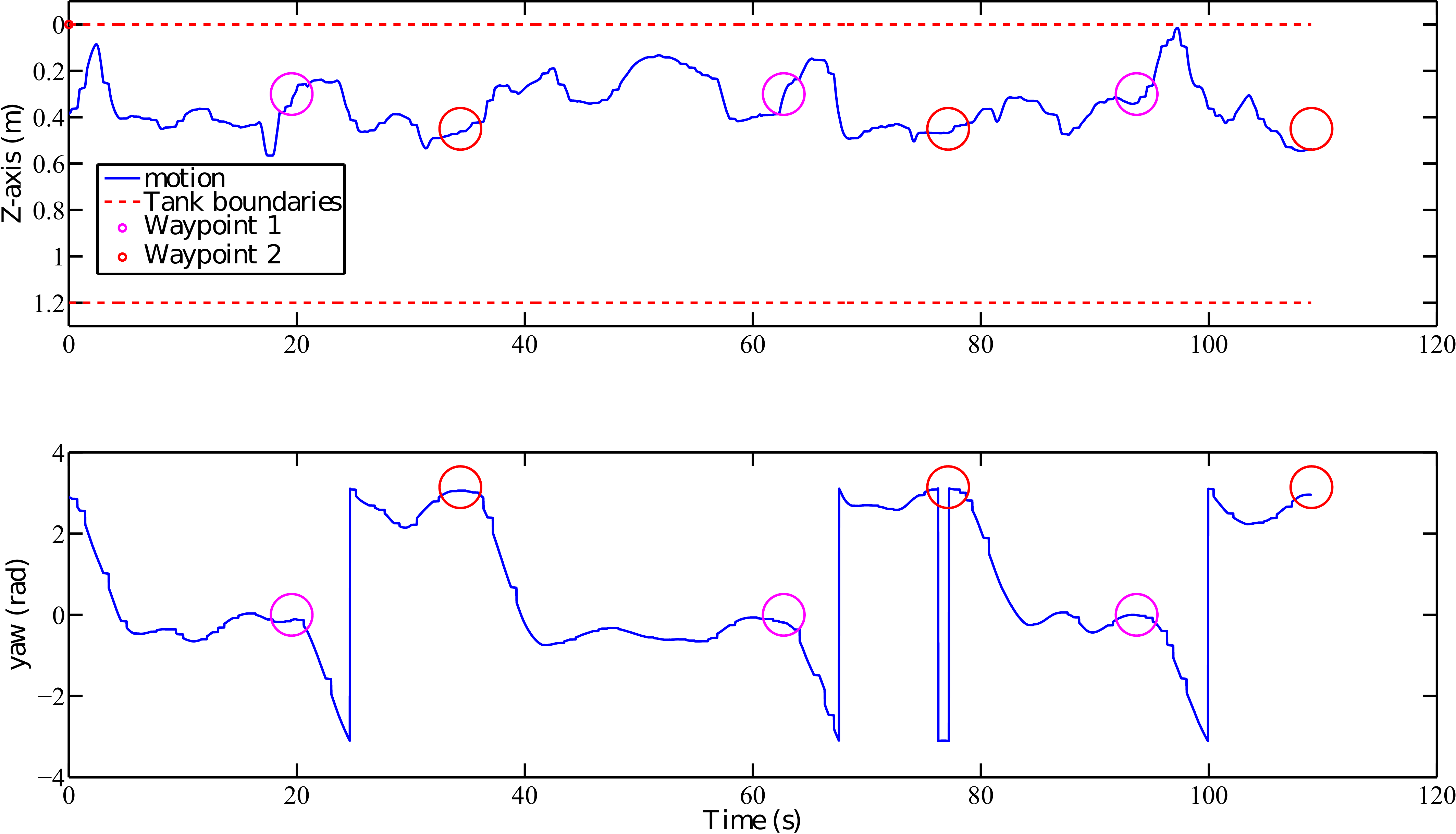}\vspace{0mm}
	\caption{2 WP tracking scenario: Vehicle vertical and angular motion\vspace{0mm} }\label{2wpzpsimotion}
\end{figure}

\begin{figure}
		\centering
	\vspace{0mm}	\includegraphics[width=3.3in]{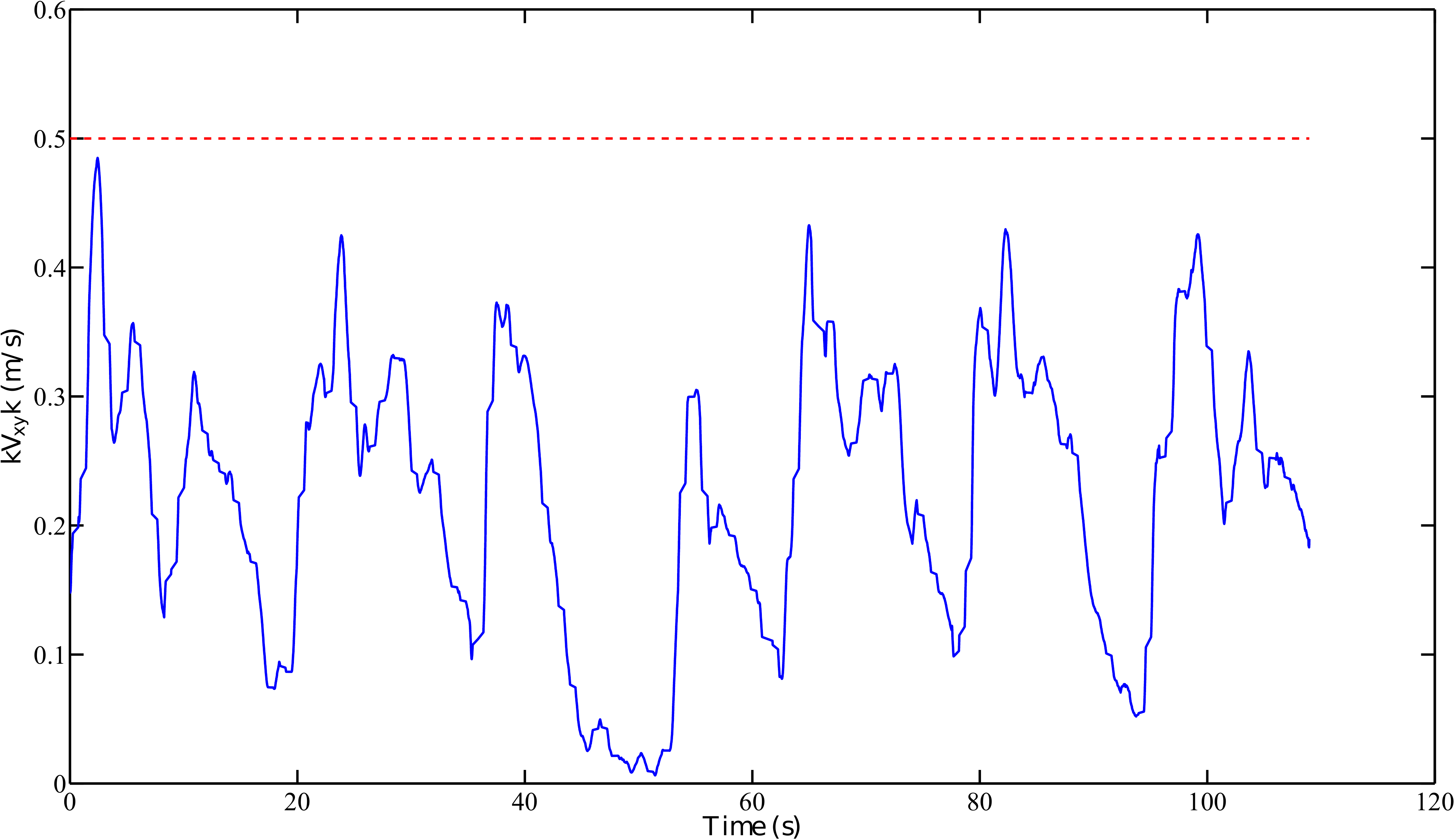}\vspace{0mm}
	\caption{2 WP tracking scenario: Vehicle body velocity norm $|u_r+v_r|$ \vspace{0mm} }\label{2wpvelnorm}
\end{figure}

\begin{figure}
		\centering
	\vspace{-2mm}	\includegraphics[width=3.3in]{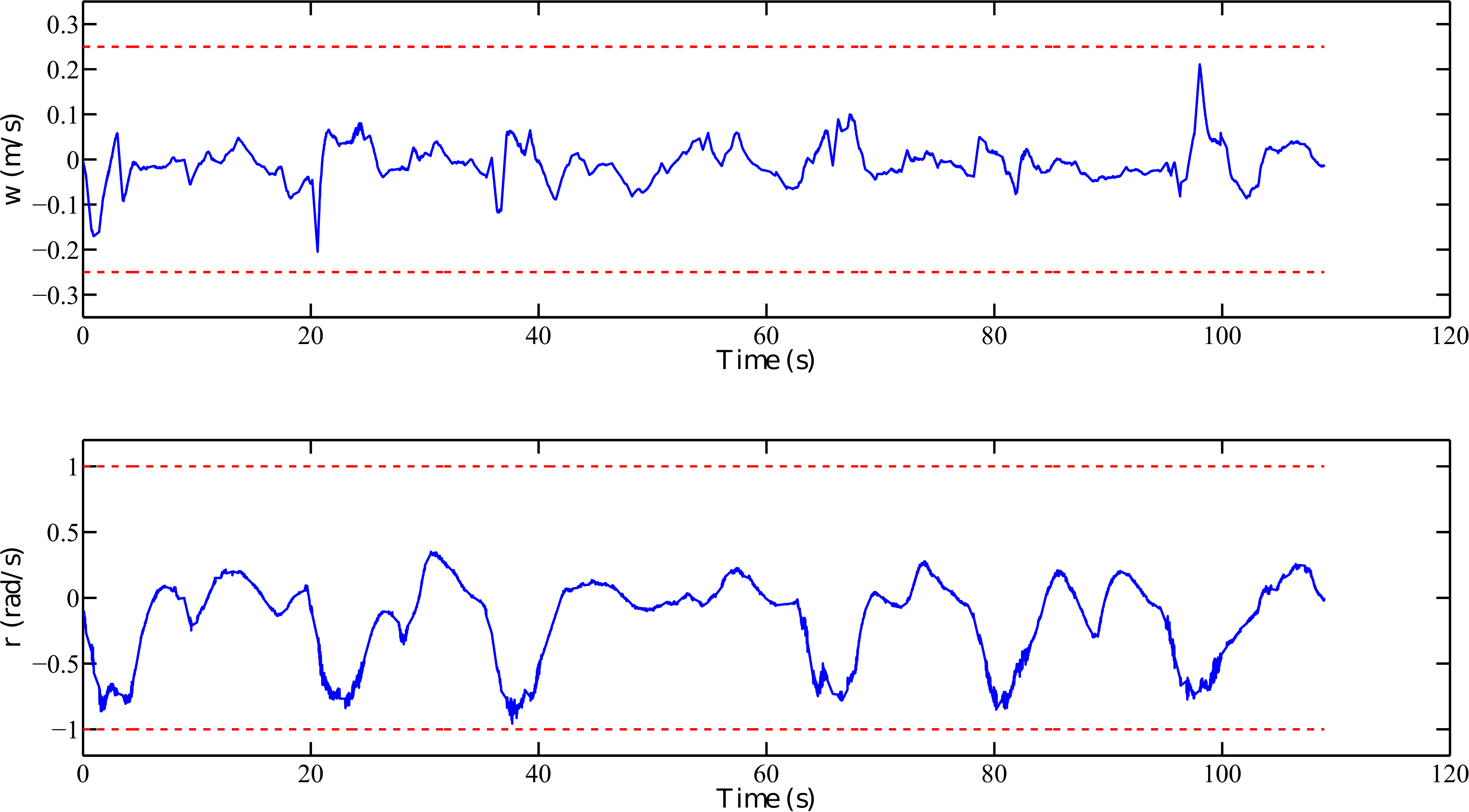}\vspace{0mm}
	\caption{2 WP tracking scenario: Vehicle heave and yaw velocities \vspace{0mm}}\label{2wpwrmotion}
\end{figure}

\begin{figure}
		\centering
	\vspace{0mm}	\includegraphics[width=3.3in]{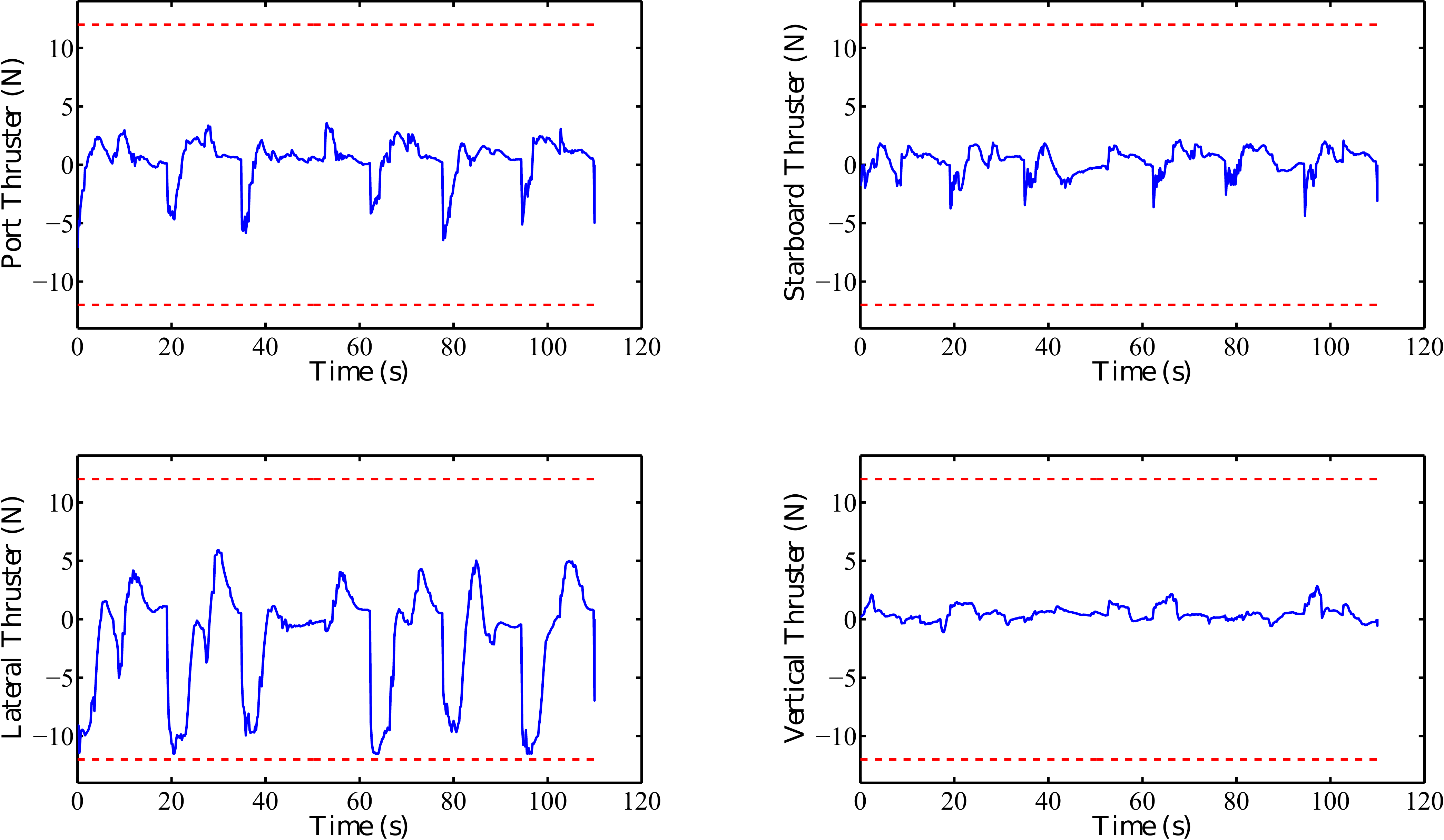}\vspace{0mm}
	\caption{2 WP tracking scenario: Thruster Commands \vspace{0mm} }\label{2wpthrusters}
\end{figure}

\textbf{Way Points Tracking Scenario:}\\
In this scenario the vehicle must travel via two waypoints which are placed at $
{\bs{\eta}^d_1 }  = \left[ {\begin{array}{*{20}c}
	{ - 1.60\,m} & { - 0.35\,m} & {0.45\,m} & {0\,rad}  \\
	\end{array}} \right],\;{\bs{\eta}^d_2 } = \left[ {\begin{array}{*{20}c}
	{1.75\,m} & {0\,m} & {0.30\,m} & {\pi \,rad}  \\
	\end{array}} \right]
$ respectively. The three consecutive trajectories of the vehicle along the horizontal plane are depicted in Fig. \ref{2wpxytrajectory}. It can be seen that the vehicle performs successfully the waypoint tracking while safely avoids the obstacles and the test tank boundaries. We also observe that in one case the vehicle traveled from the second waypoint back to first one following a different trajectory. This can be explained by the fact that the MPC found a different optimal solution at the specific time frame, due to the unmodeled dynamics of the system (e.g., tether, or vehicle dynamic parameter's uncertainties) which significantly affect the vehicle motion. The vertical and angular motion of the vehicle are depicted in Fig. \ref{2wpzpsimotion} where it can be seen that the state constraints are always satisfied. The vehicle is consider to reach each waypoint if it has entered a terminal region (i.e., spherical region of $0.3m$ and a offset of $\pm 0.15rad$) around the desired waypoint. These regions are depicted in circles in Fig. \ref{2wpxytrajectory} and \ref{2wpzpsimotion}. In Fig. \ref{2wpvelnorm} the body velocity norm in planar motion is depicted and the respective constraint is satisfied. The same stands for the heave and yaw velocities as shown in Fig. \ref{2wpwrmotion}. In Fig. \ref{2wpthrusters} the vehicle's thruster input are shown. As it can be observed, the vehicle achieved all desired waypoints while simultaneously satisfied all respective state and input constraints.


\section{CONCLUSION}\label{Sec:conclusion}
In this paper, we presented a novel Model predictive Control strategy for underwater robotic vehicles operating in a constrained workspace including obstacles. The purpose of this control scheme is to guide the vehicle towards specific way points. Various constraints such as: obstacles, workspace boundary, thruster saturation and predefined upper bound of the vehicle velocity (requirements for various underwater tasks such as seabed inspection scenario, mosaicking) are considered during the control design. Moreover, the proposed control scheme incorporates the dynamics of the vehicle and is designed in order to find optimal thrusts required for minimizing the way point tracking error. Owing to the existence of ocean currents profile in the motion dynamics, the control inputs calculated by the proposed controller may exploit the ocean currents when are in favor of the waypoint tracking mission, which results in retaining the energy consumed by the thrusters in a reduced level. Future research efforts will be devoted towards extending the proposed methodology for multiple AUVs operating in a dynamic environment including static and moving obstacles.  

{\tiny{
		\bibliography{mybibfileshahab}

\begin{thebibliography}{10}

\bibitem{griffiths2002technology}
G.~Griffiths, {\em Technology and Applications of Autonomous Underwater
  Vehicles}.
\newblock Ocean science and technology, CRC Press, 2002.

\bibitem{Fossen2011}
T.~Fossen, {\em Handbook of Marine Craft Hydrodynamics and Motion Control}.
\newblock 2011.

\bibitem{Zeng2015303}
Z.~Zeng, L.~Lian, K.~Sammut, F.~He, Y.~Tang, and A.~Lammas, ``A survey on path
  planning for persistent autonomy of autonomous underwater vehicles,'' {\em
  Ocean Engineering}, vol.~110, pp.~303--313, 2015.

\bibitem{Huynh20151144}
V.~Huynh, M.~Dunbabin, and R.~Smith, ``Predictive motion planning for auvs
  subject to strong time-varying currents and forecasting uncertainties,''
  vol.~2015-June, pp.~1144--1151, 2015.

\bibitem{Alvarez2004418}
A.~Alvarez, A.~Caiti, and R.~Onken, ``Evolutionary path planning for autonomous
  underwater vehicles in a variable ocean,'' {\em IEEE Journal of Oceanic
  Engineering}, vol.~29, no.~2, pp.~418--429, 2004.

\bibitem{Garau20095}
B.~Garau, M.~Bonet, A.~Alvarez, S.~Ruiz, and A.~Pascual, ``Path planning for
  autonomus underwater vehicles in realistic oceanic current fields:
  Application to gliders in the western mediterranean sea,'' {\em Journal of
  Maritime Research}, vol.~6, no.~2, pp.~5--21, 2009.

\bibitem{Pêtrès2007331}
C.~Pêtrès, Y.~Pailhas, P.~Patrón, Y.~Petillot, J.~Evans, and D.~Lane, ``Path
  planning for autonomous underwater vehicles,'' {\em IEEE Transactions on
  Robotics}, vol.~23, no.~2, pp.~331--341, 2007.

\bibitem{allgower2004}
F.~Allgöwer, R.~Findeisen, and Z.~Nagy, ``Nonlinear model predictive control:
  From theory to application,'' {\em the Chinese Institute of Chemical
  Engineers}, vol.~35, no.~3, pp.~299--315, 2004.

\bibitem{Medagoda2012}
L.~Medagoda and S.~Williams, ``Model predictive control of an autonomous
  underwater vehicle in an in situ estimated water current profile,'' {\em
  Program Book - OCEANS 2012 MTS/IEEE Yeosu: The Living Ocean and Coast -
  Diversity of Resources and Sustainable Activities}, 2012.

\bibitem{Fernandez201788}
D.~Fernandez and G.~Hollinger, ``Model predictive control for underwater robots
  in ocean waves,'' {\em IEEE Robotics and Automation Letters}, vol.~2, no.~1,
  pp.~88--95, 2017.

\bibitem{Jagtap2016772}
P.~Jagtap, P.~Raut, P.~Kumar, A.~Gupta, N.~Singh, and F.~Kazi, ``Control of
  autonomous underwater vehicle using reduced order model predictive control in
  three dimensional space,'' {\em IFAC-PapersOnLine}, vol.~49, no.~1,
  pp.~772--777, 2016.

\bibitem{Heshmati-Alamdari20143826}
S.~Heshmati-Alamdari, A.~Eqtami, G.~Karras, D.~Dimarogonas, and
  K.~Kyriakopoulos, ``A self-triggered visual servoing model predictive control
  scheme for under-actuated underwater robotic vehicles,'' {\em Proceedings -
  IEEE International Conference on Robotics and Automation}, pp.~3826--3831,
  2014.

\bibitem{NCOM}
{\em Naitonal HF RADAR network - surface currents}.

\bibitem{Smith20101475}
R.~Smith, Y.~Chao, P.~Li, D.~Caron, B.~Jones, and G.~Sukhatme, ``Planning and
  implementing trajectories for autonomous underwater vehicles to track
  evolving ocean processes based on predictions from a regional ocean model,''
  {\em International Journal of Robotics Research}, vol.~29, no.~12,
  pp.~1475--1497, 2010.

\bibitem{Aguiar20071092}
A.~Aguiar and A.~Pascoal, ``Dynamic positioning and way-point tracking of
  underactuated auvs in the presence of ocean currents,'' {\em International
  Journal of Control}, vol.~80, no.~7, pp.~1092--1108, 2007.

\bibitem{Fossen2}
T.~Fossen, ``Guidance and control of ocean vehicles,'' {\em Wiley, New York},
  1994.

\bibitem{Koditschek1990412}
D.~Koditschek and E.~Rimon, ``Robot navigation functions on manifolds with
  boundary,'' {\em Advances in Applied Mathematics}, vol.~11, no.~4,
  pp.~412--442, 1990.

\bibitem{Aruco2014}
S.~Garrido-Jurado, R.~M. noz Salinas, F.~Madrid-Cuevas, and
  M.~Mar\'in-Jim\'enez, ``Automatic generation and detection of highly reliable
  fiducial markers under occlusion,'' {\em Pattern Recognition}, vol.~47,
  no.~6, pp.~2280 -- 2292, 2014.

\bibitem{Marantos20161214}
P.~Marantos, Y.~Koveos, and K.~Kyriakopoulos, ``Uav state estimation using
  adaptive complementary filters,'' {\em IEEE Transactions on Control Systems
  Technology}, vol.~24, no.~4, pp.~1214--1226, 2016.

\bibitem{Quigley09}
M.~Quigley, B.~Gerkey, K.~Conley, J.~Faust, T.~Foote, J.~Leibs, E.~Berger,
  R.~Wheeler, and A.~Ng, ``Ros: an open-source robot operating system,'' in
  {\em Proc. of the IEEE Intl. Conf. on Robotics and Automation (ICRA) Workshop
  on Open Source Robotics}, (Kobe, Japan), May 2009.

\bibitem{Asouti2011232}
V.~Asouti, X.~Trompoukis, I.~Kampolis, and K.~Giannakoglou, ``Unsteady cfd
  computations using vertex-centered finite volumes for unstructured grids on
  graphics processing units,'' {\em International Journal for Numerical Methods
  in Fluids}, vol.~67, no.~2, pp.~232--246, 2011.

\end{thebibliography}
		\bibliographystyle{ieeetr}
	}}
\end{document}